\documentclass[conference]{IEEEtran}
\renewcommand\IEEEkeywordsname{Index Terms}
\usepackage{acro}
\DeclareAcronym{IMU}{
  	short = IMU ,
  	long  = Inertial Measurement Unit,
}
\DeclareAcronym{KF}{
  	short = KF ,
  	long  = Kalman Filter ,
}
\DeclareAcronym{LDA}{
  	short = LDA ,
  	long  = Linear Discriminant Analysis ,
}
\DeclareAcronym{PCA}{
  	short = PCA ,
  	long  = Principal Component Analysis ,
}
\DeclareAcronym{NB}{
  	short = NB ,
  	long  = Naive Bayes ,
}
\DeclareAcronym{UI}{
  	short = UI ,
  	long  = User Interface ,
}
\DeclareAcronym{MRC}{
  	short = MRC ,
  	long  = maximum ratio combining ,
}
\DeclareAcronym{SMNR}{
  	short = SMNR ,
  	long = Sum signal-to-noise-modulation-ratio,
}
\DeclareAcronym{WE}{
  	short = WE ,
  	long = Wavelet Entropy,
}
\DeclareAcronym{SE}{
  	short = SE ,
  	long = Spectral Entropy,
}
\DeclareAcronym{MI}{
  	short = MI ,
  	long = Motion Intensity,
}
\DeclareAcronym{PAF}{
	short = PAF,
	long = Peak asymmetry factor,
}
\DeclareAcronym{APC}{
	short = APC,
	long = Sum of Power Spectral Density,
}
\DeclareAcronym{SNR}{
	short = SNR,
	long = Signal-to-Noise Ratio,
}
\DeclareAcronym{API}{
	short = API,
	long = Application Programming Interface,
}

\usepackage{cite}
\usepackage[pdftex]{graphicx}
\ifCLASSINFOpdf
   \usepackage[pdftex]{graphicx}
   \DeclareGraphicsExtensions{.pdf,.jpeg,.png,.eps,.jpg}
\else
  \graphicspath{../images/}
   \DeclareGraphicsExtensions{.pdf,.jpeg,.png,.eps,.jpg}
\fi
\usepackage{amsmath}
\usepackage{mathtools}
\usepackage{isotope}
\usepackage{mhchem}

\usepackage{mdwmath}
\usepackage{mdwtab}
\usepackage{multirow}

\usepackage{algorithmic}

\usepackage{array}
\usepackage{epstopdf}

\ifCLASSOPTIONcompsoc
  \usepackage[caption=false,font=normalsize,labelfont=sf,textfont=sf]{subfig}
\else
  \usepackage[caption=false,font=footnotesize]{subfig}
\fi

\usepackage{color}

\usepackage{url}

\begin{document}
%
\title{Mobile Quantification and Therapy Course Tracking for Gait Rehabilitation}
\author{\IEEEauthorblockN{Javier Conte Alcaraz, Sanam Moghaddamnia, and J{\"u}rgen Peissig}
\IEEEauthorblockA{\\Leibniz Universit{\"a}t Hannover\\Institute of Communications Technology\\}
}
\maketitle
\begin{abstract}
This paper presents a novel autonomous quality metric to quantify the rehabilitations progress of subjects with knee/hip operations. The presented method supports digital analysis of human gait patterns using smartphones. The algorithm related to the autonomous metric utilizes calibrated acceleration, gyroscope and magnetometer signals from seven \acp{IMU} attached on the lower body in order to classify and generate the grading system values. The developed Android application connects the seven \acp{IMU} via Bluetooth$^{\textrm{\textregistered}}$ and performs the data acquisition and processing in real-time. In total nine features per acceleration direction and lower body joint angle are calculated and extracted in real-time to achieve a fast feedback to the user. We compare the classification accuracy and quantification capabilities of \ac{LDA}, \ac{PCA} and \ac{NB} algorithms. The presented system is able to classify patients and control subjects with an accuracy of up to 100\%. The outcomes can be saved on the device or transmitted to treating physicians for later control of the subject’s improvements and the efficiency of physiotherapy treatments in motor rehabilitation. The proposed autonomous quality metric solution bears great potential to be used and deployed to support digital healthcare and therapy.
\end{abstract}
%
\IEEEpeerreviewmaketitle
\section{Introduction}
\label{sec:Introduction}
Developing an autonomous system to assess rehabilitation progress and sport performance is of great importance in sport and clinical treatment. Usually the rehabilitation process is performed in clinics and hospitals where the patients have to accomplish different treatments and trainings.
Providing an ambulatory gait system for automatic recognition and assessment of human movement patterns is of great benefit in high-quality and flexible patient care as well as on-line follow-up of the treatment's success by healthcare providers. The use of new technologies in the health care and sport sectors is common \cite{iphone}. There are some applications to track, for example, the course of running training and the technique of the athletes \cite{AndroidAppRunnning,Wearable}. Nevertheless, there are no such applications for the rehabilitation process. Due to an aging society, increasing industrialization, and environmental factors the number of patients with knee/hip operations will grow rapidly in the coming decades \cite{centers2009racial}. Motivated by this background, the need for an autonomous quality grading system solution bears a great potential to be used and deployed to support digital healthcare and therapy. In this direction, in this paper different methods in time, frequency and time-frequency domain are utilized to extract relevant gait features. The classification accuracy of attained features is investigated using the \ac{LDA}, \ac{PCA} and \ac{NB} algorithms. A grading system is established enabling to quantify the gait rehabilitation progress.\\
The paper is organized as follows. In Section \ref{sec:MeasurementsAndMobilePlatform} the methodology for the sensor platform, data collection and the mobile platform are presented. Section \ref{sec:Feature extractionAndClassification} introduces the applied methods for feature extraction and selection as well as the classifiers used in this study. In Section \ref{sec:GradingSystem} different grading systems are explained. In Section \ref{sec:Results} the results of the classifiers and grading systems are presented. Finally, the main outcomes of this work are concluded in Section \ref{sec:Conclusion}.
\section{Measurements and Mobile platform}
\label{sec:MeasurementsAndMobilePlatform}
In this section, we introduce the measurement setup we used in this study. In particular, we present the hardware part, consisting of different motion sensors attached to the lower part of the body, and the software part, which is an Android-based application. Moreover, we discuss the way that data were collected for further processing.
\subsection{Sensor Platform}
\label{subsec:SensorPlatform}
\begin{figure} 
  \centering
  \includegraphics[width=\columnwidth]{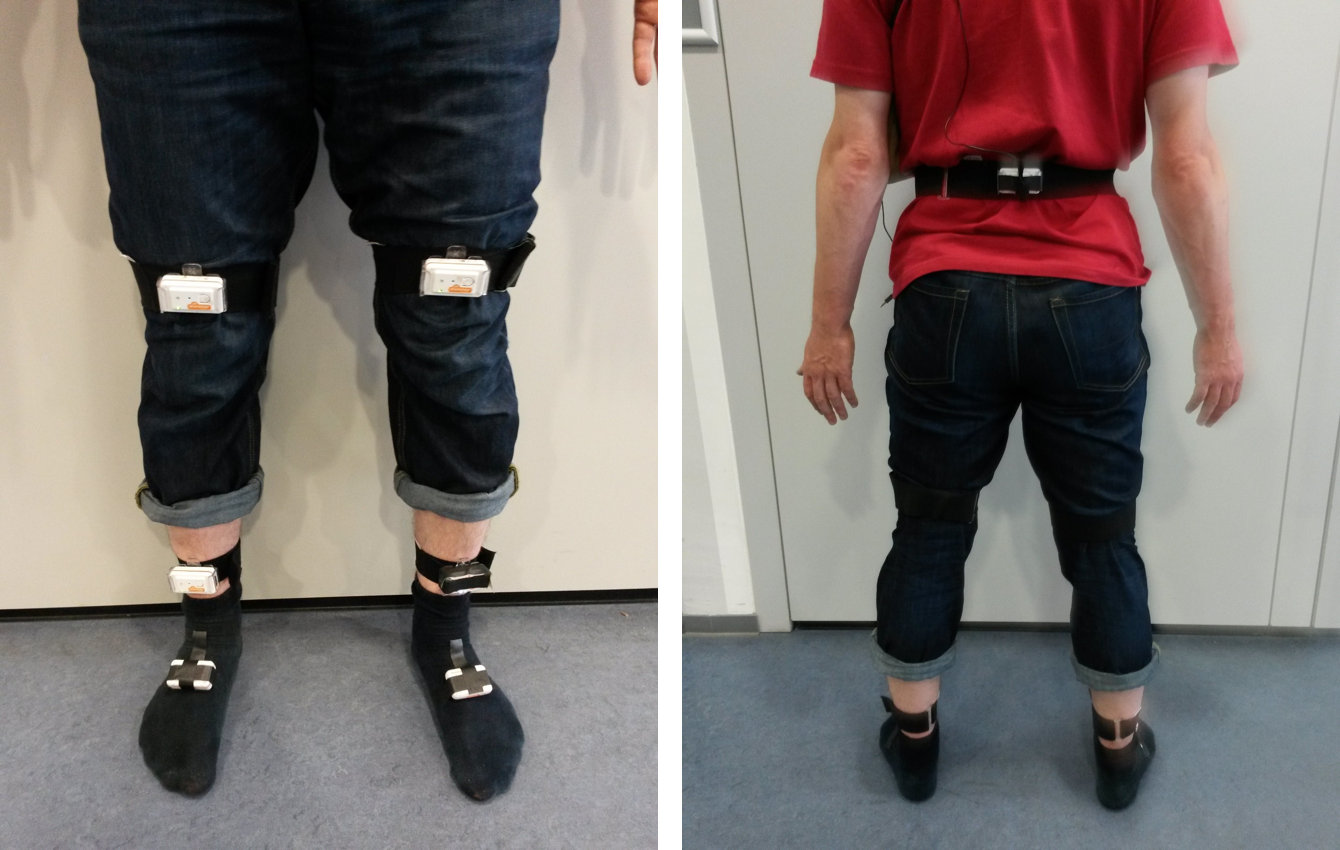}
  \caption{Sensor setup for data capturing.}
  \label{fig:SensorSetup} 
\end{figure}

We use a set of seven \acp{IMU} integrated into a sensor platform developed by the company Shimmer. In detail, we utilize the sensor platform (Shimmer 3)\cite{Shimmer}, which is capable of providing real-time motion sensing. The data acquisition is performed through Bluetooth$^{\textrm{\textregistered}}$ and transmitted to an Android application, which we developed for this purpose. The sensor units were attached to the feet, lower legs, upper legs and Pelvis, as seen in Fig. \ref{fig:SensorSetup}. To provide comparable conditions, the same sensors were attached at the same positions for each subject, who can be either patient or a healthy participant\footnote{In this paper, we use the terms {\it subject} and {\it participant} interchangeably.}. The data was captured synchronously at a sampling rate of 100 Hz. The sensor configuration is depicted in Table \ref{tab:SensorSetup}. The transmitted data was processed in real-time to estimate of the biomechanical parameters of the lower body, feature extraction, classification, and quality assessment.
\begin{table}[tb]
\centering
\caption{Sensor configuration for data acquisition}
\label{tab:SensorSetup}
\begin{tabular}{llr}
\hline
Sensor type    		& Range 			& Resolution \\
\hline
3DoF Accelerometer      	& $\pm 8$ g (m/$\text{s}^{2}$)   		& 16 bit     \\
3DoF Gyroscope       		& $\pm 500$ dps ($^{\circ}/\text{s}$)    	& 16 bit     \\
3DoF Magnetometer       	& $\pm 1.3$ Gs (100 $\mu$T)    	& 16 bit      \\
\hline
\end{tabular}
\end{table}	

\subsection{Data Collection}
\label{subsec:DataCollection}
The data was acquired from 15 patients, with an age of 65$\pm$5~years, who had knee/hip operations. For a testing purpose, the data of a reference group, consisted of 15 healthy persons with an age of 33$\pm$6~years, was also collected. After attaching the sensor to the lower body, the calibration of the sensor system was performed during a standing phase where the joint angles were set to 0$^{\circ}$. Afterwards, the measurements in a distance of 10~m straight walking were performed at the preferred speed of the patients and control subjects. Additional information such gender, age, weight and time after the operation were gathered. Seven trials were performed for each subject in each group starting with the left foot.
\subsection{Android Application}
\label{subsec:Android}
\begin{figure}[tb]
\begin{minipage}[b]{1.0\linewidth}
  \centering
  \centerline{\includegraphics[width=\linewidth]{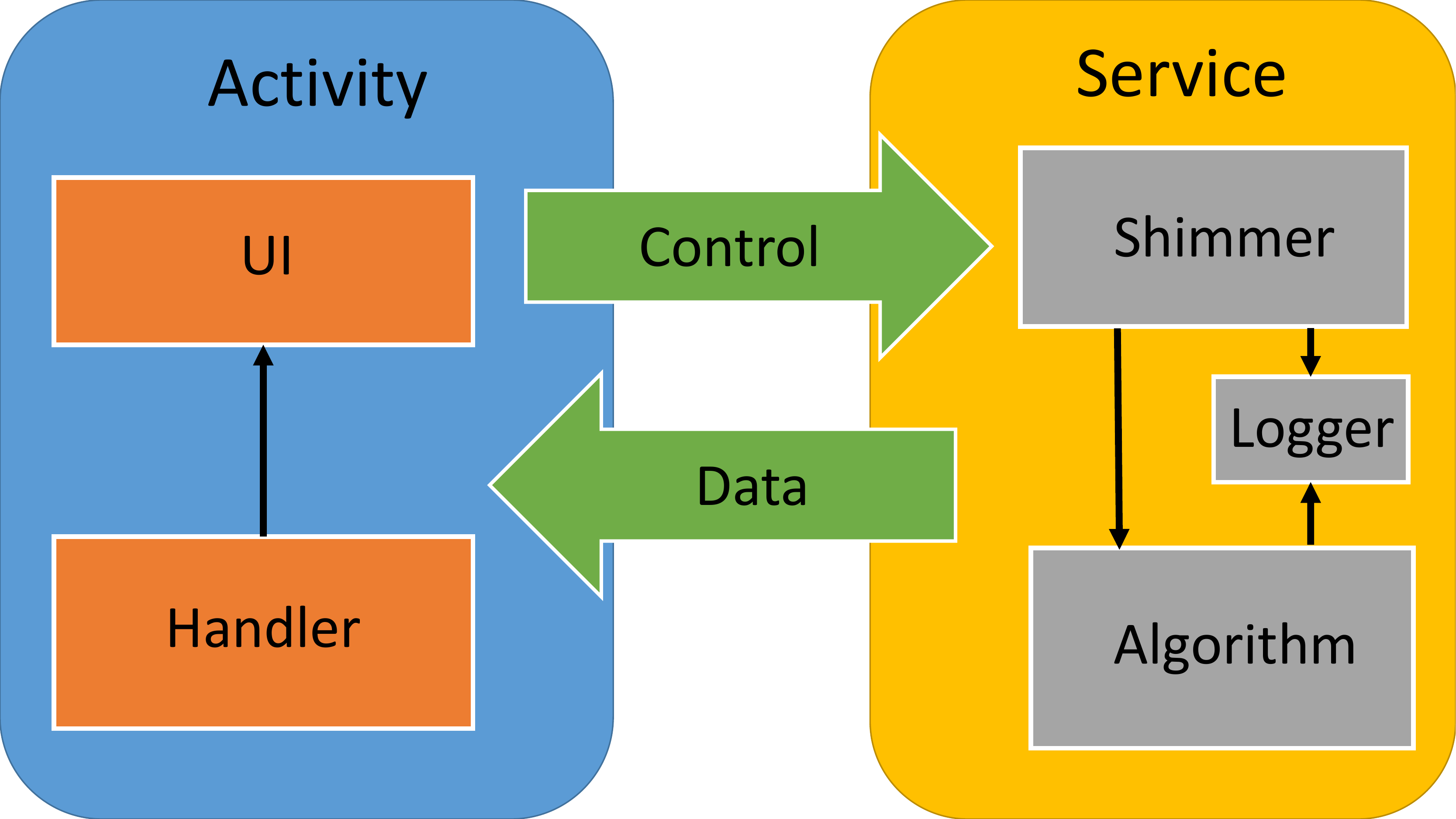}}
\end{minipage}
\caption{Design concept of the Android application}
\label{fig:service}
\end{figure}

The developed Android application was designed based on the Android programming principles and the basic features and design concepts of Shimmer \cite{Android}. The connection between the \acp{IMU} and the smartphone is performed over the Bluetooth$^{\textrm{\textregistered}}$ \ac{API} and threads. These threads establish and maintain the connection with the sensors. As depicted in Fig. \ref{fig:service}, the handler accomplishes the inter-thread communication using messages that contain the information of accelerometer data and the algorithm results. The second step is the real-time processing. In order to avoid blocking the \ac{UI} and to ensure the proper function of the application, a background service is executed, where the gait patter algorithms are running and all calculations are performed. All algorithms utilize a sample-based moving window to collect and process the accelerometer and joint angle data. The last step is the visual feedback by graphical representation of the feature values and quantification progress. It is possible to choose and display a certain set of signals and features on the \ac{UI}. Additionally, the application menu offers the facility to configure the sensors.
\section{Feature extraction and Classification}
\label{sec:Feature extractionAndClassification}
As mentioned earlier, the main purpose of this paper is to develop a sensor-based grading system for gait analysis to distinguish between healthy and unhealthy persons. To that end, in this section we describe the signal processing steps applied on the data collected from the participants. Before proceeding, we remark that all signal processing tasks are performed in a real-time sense.
\subsection{Joint Angle Estimation}
\label{subsec:JointAngleEstimation}
\begin{figure}[tb]
\centering
\includegraphics[width=\linewidth]{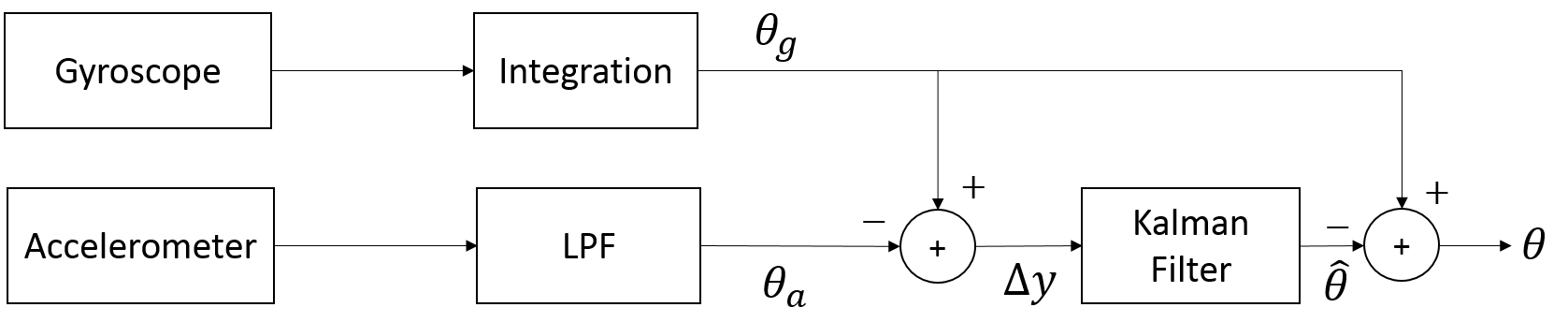}
\caption{Block diagram of the indirect Kalman Filter.}
\label{fig:BlockDiagrammKF}
\end{figure}

\begin{figure}
\centering
\includegraphics[width=\columnwidth]{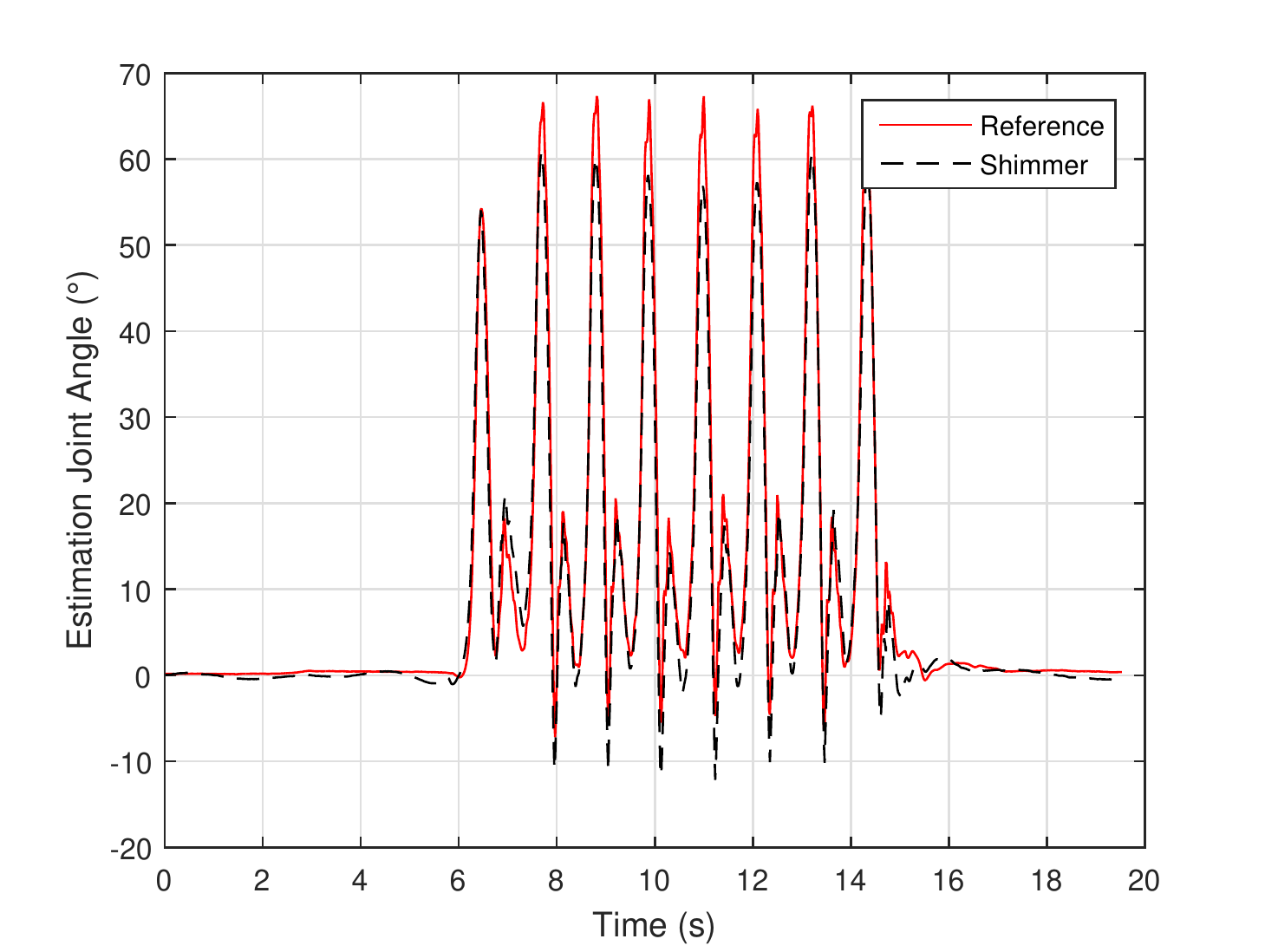}
\caption{Measured Knee joint angle compared to the reference system.}
\label{fig:ComparisonKneeShimmerReference}
\end{figure}

Joint angle estimation is known to be an important biomechanical parameter to learn the walking pattern of a person. To perform such an estimation, the sagital plane information can be accomplished using the difference between the gyroscope and acceleration angles (Fig. \ref{fig:BlockDiagrammKF}). Even with very well calibrated sensors, the joint angle obtained from the integration of angular velocity will drift after a short period of time. This drift is normally due to the temperature bias of the gyroscopes. To compensate for the drift error, an indirect \ac{KF} is applied. The state vector $\mathbf{x}_{n}$ of the \ac{KF} is defined as $\mathbf{x}_{n} = [\hat{\theta} \quad \beta]^{T}$ , where $\hat{\theta}$ and $\beta$ denote the error of the joint angle and the bias of the gyroscope measurement, respectively. The \ac{KF} uses the angle of acceleration measurements ($\theta_{a}$) as a correction to the already estimated joint angle based on the integration of the gyroscope measurement ($\theta_{g}$) \cite{Saito}. The \ac{KF} estimates the  joint angle error and subtracts it from the integrated angle to get the corrected estimation. The related state transition matrix $F$ and the measurement matrix $H$ of the \ac{KF} are given by
\begin{equation}
F=
  \begin{bmatrix}
    1 	&  T_{s} \\
    0 	& 1
  \end{bmatrix}\quad \text{and} \quad
  H=
  \begin{bmatrix}
    1 & 0
  \end{bmatrix} , 
\end{equation}
where $T_{s}$ is the sampling period of the \ac{IMU}. In Fig. \ref{fig:ComparisonKneeShimmerReference}, we demonstrate the estimation angle when a participant is walking in a straight line for a distance of 10 m. We further show the result for a data collected from another sensor platform, which we denote as reference. It is clearly shown that both results have a good matching, which means that the angle estimation algorithm discussed herein is independent of the hardware platform used for data collection. 

\subsection{Feature Extraction and Selection}
\label{subsec:FeatureExtraction}
Now having the joint angle estimations, we perform the feature extraction step where we utilize the acceleration signals from the sensors and the \ac{KF} output signals. Prior to that, a rotation of the acceleration signals from the sensor frame S into the global frame G is required \cite{kuipers}. In principle, a three-dimensional vector can be rotated by a quaternion $\mathbf{q}$, using the following relationship \cite{Madgwick}:
\begin{equation}
\ce{^{G}_{}\mathbf{a}} = \ce{^{G}_{S}\mathbf{q}} \otimes \ce{^{S}_{}\mathbf{a}} \otimes \ce{^{G}_{S}\mathbf{q}^\ast} ,
\label{eq:QuaternionRotation}
\end{equation}
where $\otimes$ and $\ast$ denote the quaternion multiplication and the conjugate operation, respectively. The quaternions are calculated using an optimised gradient-descent algorithm described in \cite{Madgwick}. Above, $\ce{^{S}_{}\mathbf{a}}$ and $\ce{^{G}_{}\mathbf{a}}$ are the same acceleration described in the sensor frame and the global frame, respectively. After removing the gravity component, the acceleration signal $\ce{^{S}_{}\mathbf{a}}$ is filtered using a Butterworth low-pass filter with a cuttoff frequency of 7 Hz to reduce the noise and the high frequency components \cite{Frequenz}. 

In order to develop a reliable grading system and an accurate classification, different feature extraction methods are considered in the literature. While time domain features specify the signal shape and statistics as reported in \cite{1sanam,JCO}, frequency domain features are mainly based on the periodic structure of the data, such as spectral properties and entropy \cite{5759000}. In this paper, the main concern is made on features that can be extracted from walking data as listed in Table \ref{tab:SignalFeatures}. As we computed the features for each accelerometer sensor, direction and joint angle, this procedure results in $[7(\text{sensors}) * 3(\text{directions}) + 6 (\text{joint angles})]*9(\text{features per each signal})=243$ total features, where 'directions' refer to the $(x,y,z)$ Cartesian coordinates of the global frame.
\begin{table*}[!tb]	
\caption{Signal Features}
\label{tab:SignalFeatures}
\begin{tabular}{lll}
\hline
Name    		& Type 			& Description \\
\hline
	 \ac{MI}		& Time analysis 			& Characterization of the movement intensity	\\
	 \ac{PAF}		& Time analysis				& Measure of the signal symmetry	\\
	 Step period	& Time analysis				& Time between steps	\\
	 Stride period 	& Time analysis				& Time between two steps of the same side	\\
	 Regularity 	& Time analysis				& Characterizes the signal rhythmic and periodicity	\\
	 \ac{APC} 		& Frequency analysis		& Mechanical power of the signal\cite{JCO}	\\
	 \ac{SE} 		& Frequency analysis		& Derived from information theory, a measure of the uncertainty of a signal\cite{Haikin}\\
	 \ac{SMNR} 		& Frequency analysis		& Characterizes the random variation relative to the periodicity \cite{SanamSMNR}	\\
	 \ac{WE} 		& Time-Frequency analysis	& Measure of the signal distortion and provide knowledge on the dynamic process		\\
\hline
\end{tabular}
\end{table*}
The features mentioned above can be treated as significant if they allow for a differentiation between normal and abnormal gait. The utilization of the paired t-test allows to distinguish the relevance of the features \cite{montgomery2010applied}. In this part of the study two groups, each consisting of eight subjects, are involved. The first group includes subjects with normal gait patterns, while the second group includes subjects with knee and hip operations. The above-mentioned features are calculated for all the subjects and those features with p-values less than $0.05$ are considered significant \cite{montgomery2010applied}. Subsequently, the optimal feature set is obtained using \ac{SNR} ranking. The \ac{SNR} for each feature can be obtained from \cite{Yang}:
\begin{equation}
SNR = \frac{\mu_1 - \mu_2}{\sigma_1 + \sigma_2}  ,
\label{eq:SNR}
\end{equation}
where $\mu_1$ and $\mu_2$ are the mean of features for subjects with the normal and abnormal walk, respectively. $\sigma_1$ and $\sigma_2$ are the corresponding standard derivations. Intuitively, a more efficient classification is expected to be achieved using features of higher \ac{SNR} values. Subsequently, after several trails we selected a total of 26 features with high \acp{SNR} to be used for the further investigations in this paper.  
\subsection{Classification}
\label{subsec:Classification}
Based on the final feature set, a classification approach is applied such that the goal is to classify each participant has either a normal or an abnormal gait pattern. In this context, supervised and unsupervised classification schemes are considered. As a result of the theorem of ``No Free Lunch''\cite{Duda}, there is no optimal classifier. Alternatively, three different classifiers are evaluated in this study: \ac{LDA}, \ac{PCA} and \ac{NB}. The final feature set is used for the classification of eight patients and eight control subjects in the training phase to define the decision boundaries for the group separation. Once the decision boundaries are specified, the test data from seven patients and seven control subject is used to verify the classification efficiency.
\begin{table}[tb]	
\centering
\caption{Classification results}
\label{tab:ClassificationResults}
\begin{tabular}{cccc}
\multicolumn{4}{c}{} \\
\hline
Classifier    		& Accuracy 			& Sensitivity & Specificity \\
\hline
\ac{PCA} 	& 85.7\%  	& 85.7\% 	& 85.7\%\\
\ac{LDA} 	& 100\%  	& 100\% 	& 100\%	\\
\ac{NB} 	& 100\%  	& 100\% 	& 100\%	\\
\hline
\end{tabular}
\end{table}	
\section{Grading System}
\label{sec:GradingSystem}
After classifying the gait pattern into normal or abnormal, in this section a grading system is proposed to quantify how normal or abnormal each patient is. This grading system can be used to track the rehabilitation progress of patients. Based on the final feature set, three different grading systems for the rehabilitation tracking are proposed. The basic concept of grading system is based on the \ac{MRC}, performed by 
\begin{equation}
G = \sum^{N}_{i} F_{i}W_{i}  ,
\end{equation}
where $N = 26$ is the number of the selected features, $F_{i}$ is the feature value, and $W_{i}$ is the weight of the contribution of each feature \cite{SanamGrading}. The first grading system utilizes the \ac{SNR} values as the weight vector. The two other grading systems are based on the separation principal used by \ac{LDA} and \ac{PCA}. Here, the grading value G, is calculated using the eigen vectors of the \ac{LDA}/\ac{PCA} as the weight values instead of the \ac{SNR} values. \ac{LDA} and \ac{PCA} mainly differ in the related orthogonal basis used for. \ac{LDA} looks for a feature space on which to project all data, such that the samples are maximally separated. \ac{PCA} finds a feature space based on the festures's deviation from the global mean in the primary directions of variation in feature space \cite{Theodoridis}.
\begin{figure}[tb]
\centering
\includegraphics[keepaspectratio, width=\columnwidth , scale=0.8]{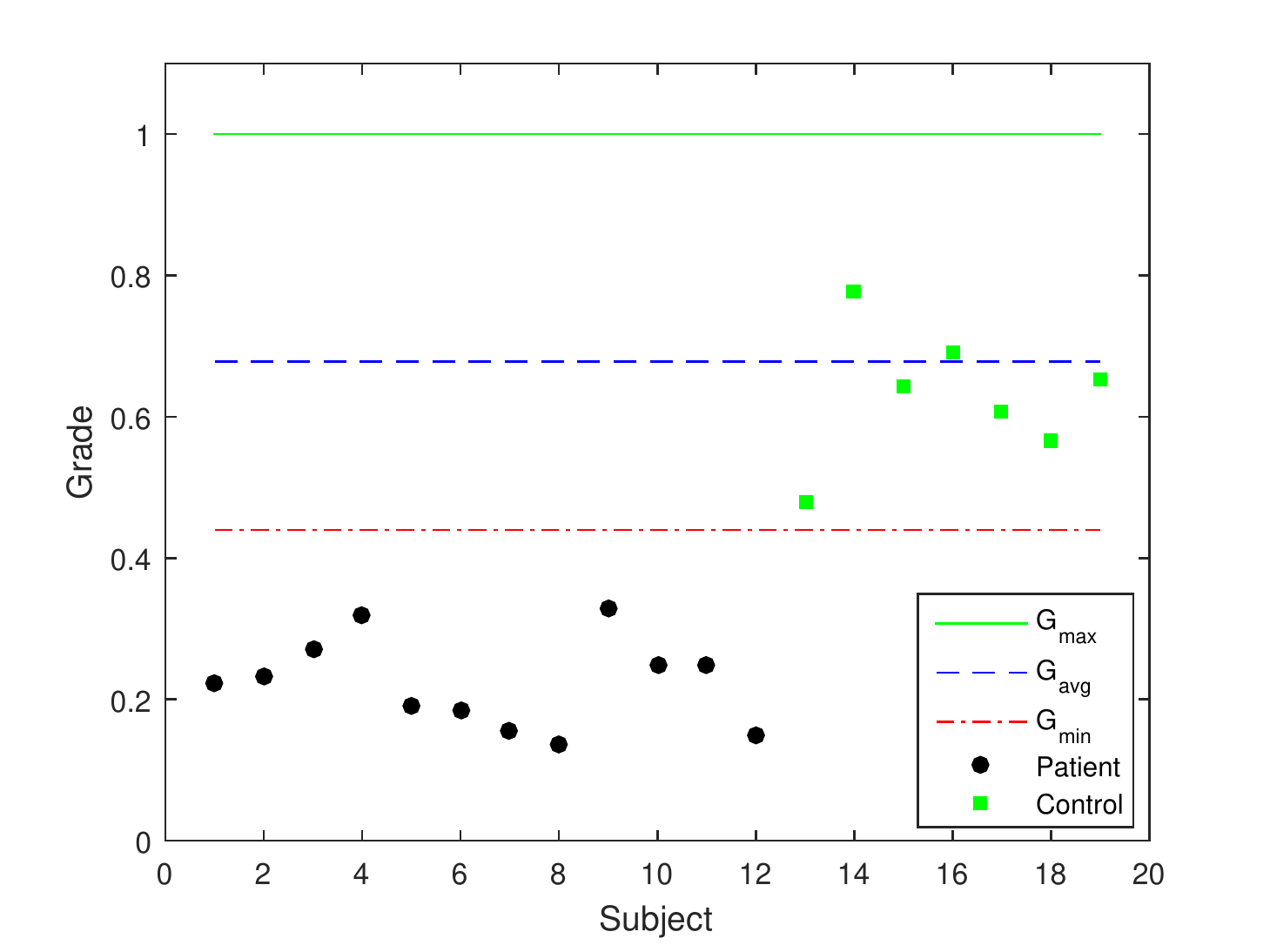}
\caption{Efficiency of the grading system based on \ac{SNR}}
\label{fig:SNR_grading}
\end{figure}
\begin{figure}[tb]
\centering
\includegraphics[keepaspectratio, width=\columnwidth , scale=0.8]{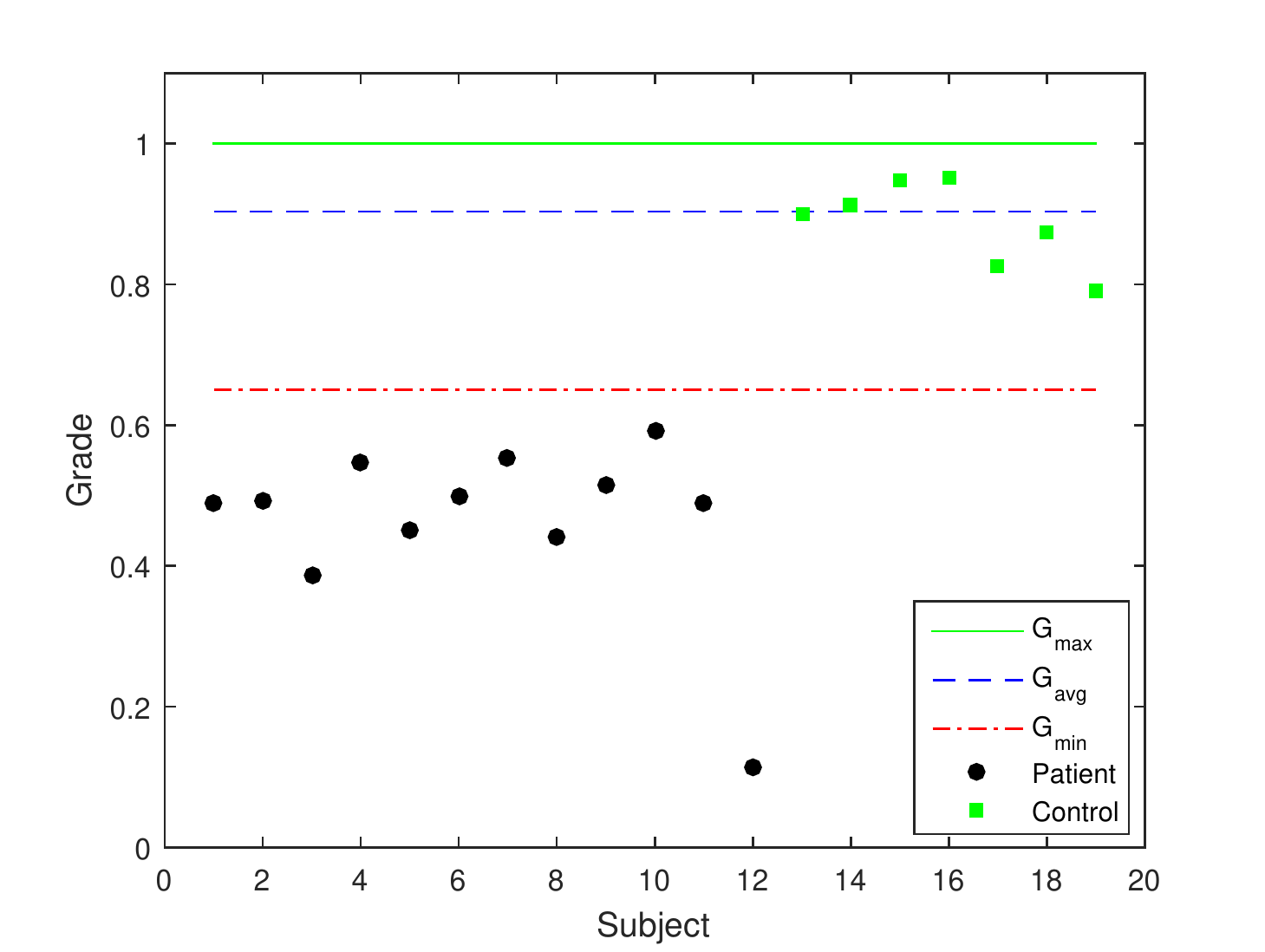}
\caption{Efficiency of the grading system based on \ac{LDA}}
\label{fig:LDA_grading}
\end{figure}
\begin{figure}[tb]
\centering
\includegraphics[keepaspectratio, width=\columnwidth , scale=0.8]{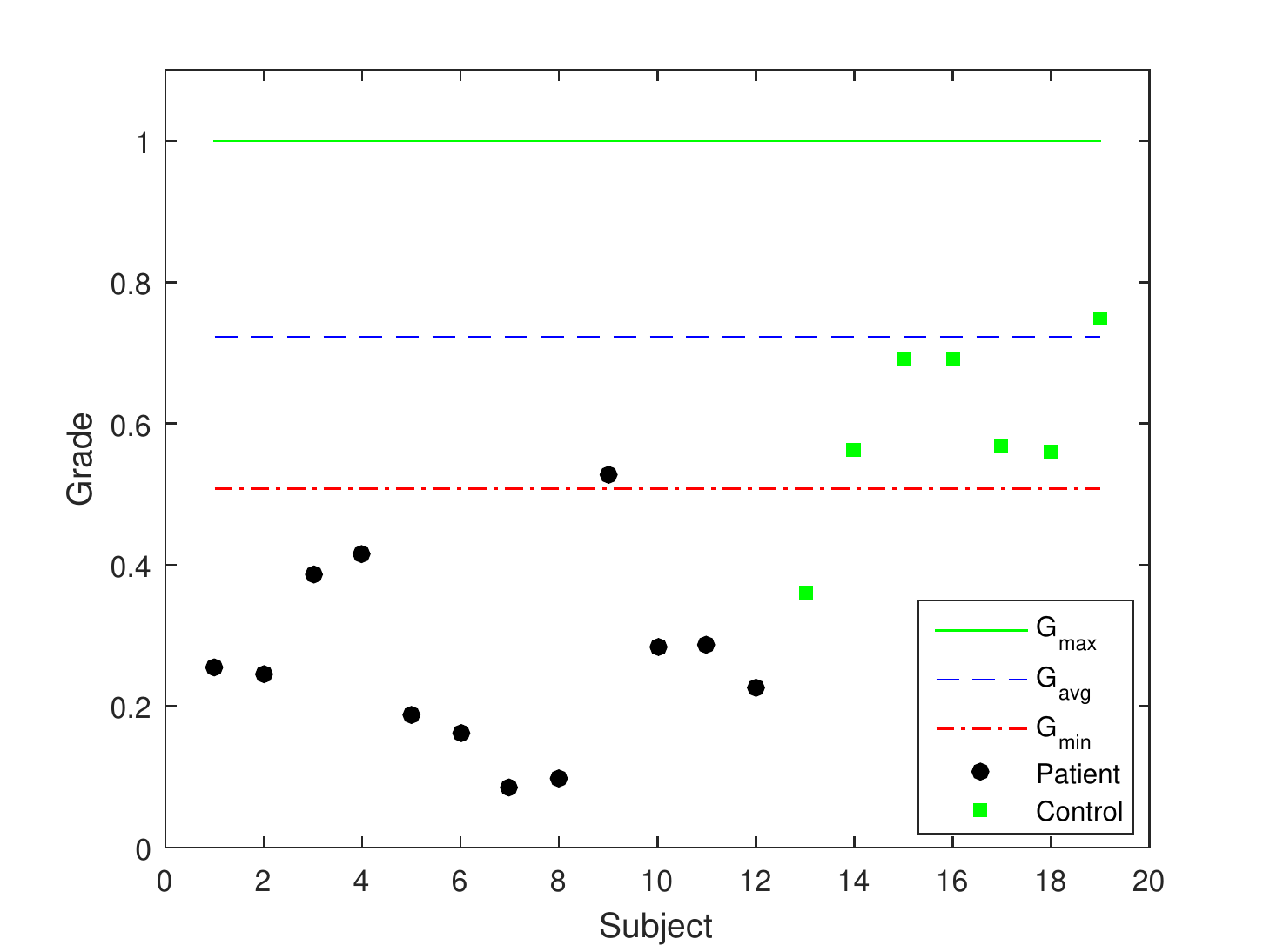}
\caption{Efficiency of the grading system based on \ac{PCA}}
\label{fig:PCA_grading}
\end{figure}


\section{Results}
\label{sec:Results}
\begin{figure}[tb]
\centering
\includegraphics[keepaspectratio, width=\columnwidth ]{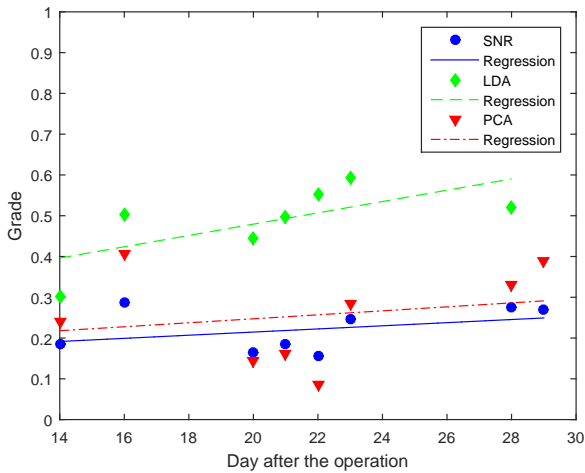}
\caption{Correlation between the grades and the time after the operation.}
\label{fig:CorrelationGradingSystem}
\end{figure}

In this section, we validate the performance of the proposed grading systems. For this purpose, we first have a training phase in which the grading boundaries are defined in terms of maximum, minimum and average values, i.e., $G_{max}$, $G_{min}$, and $G_{avg}$ as shown in Fig. \ref{fig:SNR_grading} to \ref{fig:PCA_grading}. Next, a test phase takes place in which we evaluate the performance of the classification and the grading systems. Table \ref{tab:ClassificationResults} compares the three considered classifiers in terms of the accuracy, sensitivity, and specificity. These results are also reflected in Fig. \ref{fig:SNR_grading} to \ref{fig:PCA_grading}, where the participants with grades between $G_{max}$ and $G_{min}$ are considered as healthy (control) ones. Finally, Fig. \ref{fig:CorrelationGradingSystem} depicts the time correlation between the grades and the days after the operation. It is clearly seen that the grades increase with the days after the operation with all considered grading systems. This means that the proposed grading systems are able to objectively quantify the gait performance and the  rehabilitation progress based on the final features set derived from the gait analysis. Moreover, it is possible to provide a numerically performance comparison of patients. From Fig. \ref{fig:SNR_grading} to \ref{fig:CorrelationGradingSystem}, we can conclude that the LDA-based grading system is preferred for two reasons: First, it has a high classification accuracy, and second, it has the highest grading-time correlation, and hence it has a better tracking capability. Additionally, we can obtain an individual profile by calculating the grade of each feature separately. 

\section{Conclusion}
\label{sec:Conclusion}
The main objectives investigated in this work are twofold: First, differentiation between patient and control group to track the therapy course. Second, objective quantification of the subject performance. For both of these tasks a high system accuracy is required. Therefore, we proposed an application for \ac{IMU} platforms which is able to precisely quantify the rehabilitation progress associated with knee/hip operations and to objectively classify between the patient and control groups. Our proposed approach allows a mobile and comfortable therapy. The assessment of therapy methods, the verification of the accuracy for the proposed grading system and an extended statistical analysis is the subject of future investigations. The final goal is to come up with a medically approved system that can be implemented into everyday clinical practice.
\section*{Acknowledgment}
\label{sec:Acknowledgment}
This work is carried out in the scope of the Central Innovation Programme SME research project in cooperation with the company MediTECH Electronic GmbH and the Institute of Sports Science of Leibniz Universit{\"a}t Hannover.

\bibliographystyle{IEEEtran}
\bibliography{myDSPBIB}%
\end{document}